\documentclass{svproc}
\usepackage{xcolor}

\usepackage{url}

\usepackage{graphicx}

\graphicspath{{images/}}

\usepackage{booktabs}

\usepackage{multirow}

\begin{document}
\mainmatter              % start of a contribution
\title{Exploiting the relationship between visual and textual features in social networks for image classification with zero-shot deep learning}
\titlerunning{Exploiting the relationship between visual and textual data...}  % abbreviated title (for running head)
%                                     also used for the TOC unless
%                                     \toctitle is used
%
\author{Luis Lucas\inst{1} \and David Tomás\inst{1} \and Jose Garcia-Rodriguez\inst{1}}
\authorrunning{Lucas et al.} % abbreviated author list (for running head)
%
%%%% list of authors for the TOC (use if author list has to be modified)
\tocauthor{Author1, Author2, and Author3}
\institute{Institute of Informatics Research, University of Alicante, Alicante, Spain,\\
\email{(luis.lucas,dtd,jgr)@ua.es},\\ %WWW home page:
%\texttt{http://xxx}
}

\maketitle              % typeset the title of the contribution

\begin{abstract}

One of the main issues related to unsupervised machine learning is the cost of processing and extracting useful information from large datasets. In this work, we propose a classifier ensemble based on the transferable learning capabilities of the CLIP neural network architecture in multimodal environments (image and text) from social media. For this purpose, we used the InstaNY100K dataset and proposed a validation approach based on sampling techniques. Our experiments, based on image classification tasks according to the labels of the Places dataset, are performed by first considering only the visual part, and then adding the associated texts as support. The results obtained demonstrated that trained neural networks such as CLIP can be successfully applied to image classification with little fine-tuning, and considering the associated texts to the images can help to improve the accuracy depending on the goal. The results demonstrated what seems to be a promising research direction.

\end{abstract}

\section{Introduction} \label{intro}

The high cost of processing and extracting useful information for its application in different areas of interest is one of the major issues related to unsupervised machine learning. Models based on neural networks need to be trained with similar data to the ones they are going to predict or intend to model. Furthermore, depending on the type of data, its characteristics, dataset size, and other factors, one architecture could be more convenient than another. Usually, the cost of adequate training consumes too much time and resources that are not available to most organizations.

The release of new models based on neural networks, which have been trained with huge amounts of data, allows transferring their knowledge to different tasks and areas. In this paper, the goal is to classify images and text from social media to allow obtaining specific and useful information in different areas of knowledge of social sciences such as economics or sociology.

In this work, we will first use CLIP (Contrastive Language-Image Pre-Training) \cite{Radford2021} transformer to classify images of a dataset obtained from social networks containing picture-text pairs obtained with a pre-trained neural network into 205 labels corresponding to places or scenes. We will then refine such classification with the associated texts to check and compare the obtained accuracy. For this purpose, we will use a sampling-based validation since the dataset used is unlabeled.

The remainder of the paper is structured as follows: Section \ref{related} shows related work;  Section \ref{system} outlines the pieces of the system that we will use in our experiments; Section \ref{setup} describes the image classification workflow, firstly using only the images and secondly by refining the classification with the associated texts from the dataset; Section  \ref{evaluation} shows the validation process and the results obtained; finally, in Section \ref{future} we discuss our conclusions and suggest future work.

\section{Related work} \label{related}

There are numerous works related to multimodal machine learning as well as the classification into unsupervised or semi-supervised techniques. The following paragraphs highlight some of the most relevant works in this area.

\subsection{Multimodal machine learning}

In this field, two evolutions of existing models, VL-T5 and VL-BART for vision and text to generate text are proposed in \cite{Cho2021}. 

In \cite{Kumar2020e}, they proposed a multimodal system that uses both the text and visual content on Twitter to classify information during emergencies. For this purpose, they use LSTM and VGG-16 for tweet texts and images, respectively. Along the same line, other proposals \cite{Kumar2020h} defined models to detect disasters in cultural heritage from social media information, but in this case only from images. In a completely different field, \cite{Cai2020} used the relationship between images and their associated texts in social media to try to detect sarcasm.  To this end, they build a single vector from the features of the image and the embeddings of the text.

Finally, in \cite{Embeddings2016} it is proposed a common image-text embedding space for training a bidirectional network. They provide retrieval results on Flickr30K and MSCOCO datasets that considerably exceed the state-of-the-art.

\subsection{Classification into unsupervised or semi-supervised techniques}

In \cite{Guillaumin2010} the authors presented a semi-supervised proposal that can train a classifier from a very small labeled subset concerning the dataset.  On the other hand, in \cite{Wang2017} they proposed a framework to automatically label images from social media with the help of their associated tags. Moreover, they test the efficiency of their methods on real social media datasets such as MIR Flickr and NUS-WIDE, with promising results. 

The following work \cite{Lin2015} proposes a dual model for classifying large datasets based on k-nearest neighbors techniques to separate clean data from noisy data, to make two different training datasets. 

On the other hand, \cite{Jiang2019} proposes a novel transferable contrastive network for generalized zero-shot learning. They establish a relationship between the classes the network has been trained on and other new classes for a different classification.

\section{Description of the system} \label{system}

The proposed system aims to classify images in a noisy environment such as a social media network not having been trained with any of the images to be classified.

To accomplish this classification task without the requirement of ad-hoc training on the target dataset, we need a pre-trained model with transferable learning capability.

The state of the art suggests using pre-trained models, eliminating the last classification layer to obtain the embedding vectors where all the features learned by the model are located. This process is well documented in recent works such as \cite{Wang2017} and \cite{Zhang2019}.

\begin{figure}[htp]
    \centering
    \includegraphics[width=\textwidth]{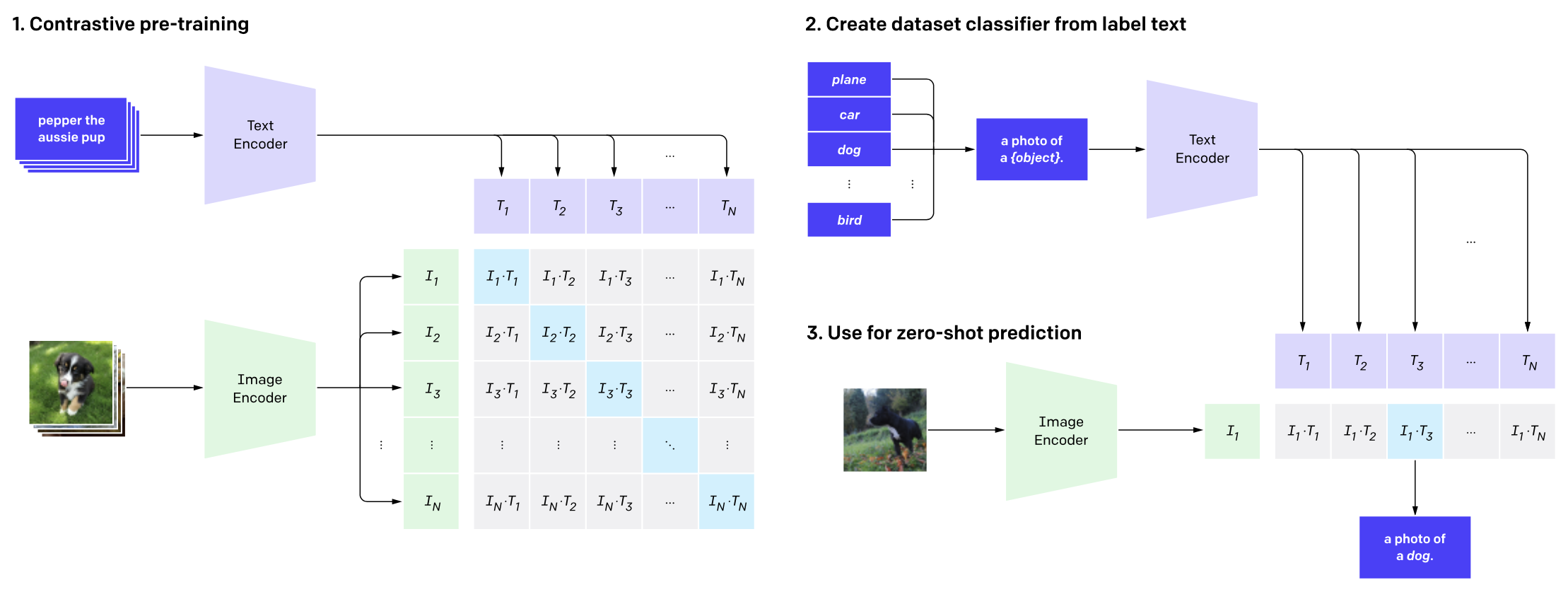}
    \caption{CLIP pipeline for zero-shot prediction.}
    \label{fig:clip}
\end{figure}

The task of classification requires training. To avoid this, it is necessary to find a common representation space that allows us to compare texts and images. In this work, this task is accomplished by using CLIP (Contrastive Language-Image Pre-Training) \cite{Radford2021}. The architecture of this model is shown in Figure \ref{fig:clip}. It consists of a pre-training neural network on a dataset of 400 million (image, text) pairs collected from the internet, learning an image representation from scratch,  that can perform the task of predicting pairs of caption and related images. After a pre-training stage, natural language is used to reference learned visual concepts (or describe new ones) enabling the zero-shot transfer of the model to downstream tasks. This model transfers non-trivially to many tasks and is often competitive with a fully supervised baseline without the need for any specific dataset training. For instance, it matches the accuracy of the original ResNet-50 on ImageNet zero-shot without needing to use any of the 1.28 million training examples it was trained on.

Our first goal is to classify and select images from a raw dataset, without the need for previous training. This classification will also help us to discard images that are not of interest to the study since CLIP provides the link between image and text that will allow the classification according to a defined taxonomy.

Finally, we will use the text associated with each of the images to help the classification process. We will exploit the common space provided by CLIP to evaluate the similarity between the embedding vector of the image and the associated text. This text, coming from social media, is noisy. Therefore, we propose a conditional ensemble for classification, adjusted according to our validation tests, acting as a multimodal constractive conditional learning, which has already been used in other works applied to specific fields such as radio diagnostics in medical imaging \cite{Zhang2020}.

\section{Experimental setup}
\label{setup}

In this work, the InstaCities1M Dataset \cite{Gomez2019} was used for the evaluation. This dataset contains social media images with the associated text. It comprises Instagram images located at one of the 10 most populated English-speaking cities all over the world, 100K images for each city, which makes a total of 1M images split in 800K for training, 50K for validation, and 150K for testing.

From the 10 cities, we decided to choose New York, as it is a well-known city, easily identifiable in photos by most people. Therefore, our dataset will consist of 100,000 pairs of images and their associated text. The images in the dataset were retrieved by searching the name of the city in the associated text or tag. The provided images include a variety of scenes, such as indoor and outdoor, people and objects, concrete actions and posters, etc. This variability is useful for our experimentation because we will use CLIP not only to obtain useful information when performing a classification but also as a second phase of refinement in the construction of a clean and useful dataset for a specific task.

Regarding the classification task, we chose a subset of the classes proposed in the MIT Places dataset \cite{Zhou2014}. It contains 205 types of places, which can be both indoor and outdoor, such as banquet hall, bar, or baseball field.

\subsection{Image classification with CLIP}

CLIP has been trained with a dataset of 400 million (image, text) pairs collected from a variety of publicly available sources on the Internet. We queried CLIP with the corresponding label of the class to retrieve related images (e.g. ``skyscraper''). To perform the classification, we have added a final softmax layer to the results obtained by CLIP. The procedure consists of asking the model for each image, how much it matches the 205 Places tags. The softmax will help us to evaluate the result. 

Our first attempt to classify the dataset with the Places labels yields the following results: using a threshold probability of 0.5, the model has classified 15,308 images out of the 100,000 in the dataset.

To analyze the results for the first experimentation, we will test a subset composed of the first 1000 images. In this case, taking a threshold of 0.5 as the probability that the image matches a particular label, we have managed to successfully label 174 out of 1000 images. This does not necessarily mean that the labeling has failed. Keep in mind that the dataset contains images that can be anything and that match none of the tags we used. One of the objectives is to clean the dataset and keep only the images of interest for us.

We must take into account that the sum of all the probabilities of each tag is 1, therefore, if in the list of tags we have two or more similar ones, the probability could be diluted. For example, in the Places tags, we have the ``art gallery'' and the  ``art studio'' tags. An image could give a probability of 0.3 to one tag and 0.3 to the other.

Table \ref{table:table1} shows the percentage of successfully tagged images, depending on the threshold probability chosen as a minimum to be able to classify an image:

\begin{table}
\centering
\caption{Percentage of images classified according to the probability threshold selected.}
\begin{tabular}{cc}
\toprule
threshold probability & \% of images classified \\
\midrule
0.5                   & 17,5                    \\
0.4                   & 25                      \\
0.3                   & 37’6                    \\
0.2                   & 56,4                    \\
0.1                   & 88,8                    \\
\bottomrule
\end{tabular}
\label{table:table1}
\end{table}

\begin{figure}[htp]
    \centering
    \includegraphics[width=8cm]{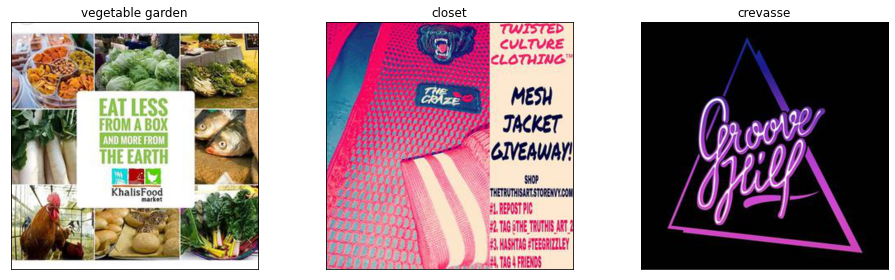}
    \caption{Images incorrectly classified.}
    \label{fig:incorrect}
\end{figure}

Examples of misclassified images are shown in Figure \ref{fig:incorrect}. Some of these errors are due in part to the fact that either, there is no appropriate text that corresponds to the image on the labels of the places, or that it is a sign or poster that may confuse CLIP, as in the last image corresponding to a luminous sign of a club. This case is interesting because it labels as ``crevasse'' a sign where it indicates ``Groove Hill''.

In most of the images not successfully classified, the label with the highest probability is the correct one, but its value does not exceed the threshold set at 0.5. Having a dataset with so much noise and with such a variety of images, if we decrease the threshold further, we get misclassified images, and if we increase it we discard too many images. 

We are going to try to solve this problem by following the OpenAI recommendations for CLIP. For the new classification, we will modify the Places classes so that these are a text more similar to the one the CLIP system has been trained with. Some examples of adaptation are:
\begin{itemize}
  \item"outdoor cathedral" → "A photo of the outdoor of a cathedral".
  \item"outdoor apartment building" → "A photo of the outdoor of an apartment building".
  \item"bakery shop" → "A photo of a bakery shop".
\end{itemize}

After the classification process with these modified versions of labels, we have been able to verify that more images are classified with a threshold of 0.5 probability, achieving 18.2\% of classified images, and if we decrease the threshold to 0.4, we also obtain a slight increase in the number of classified images reaching up to 26.40\%. Thus we have kept this adaptation to a more natural language of the labels of the Places dataset when assessing and evaluating our research.

\subsection{Improving the classification with the text associated with the images.}

All the images in our dataset have an associated text. This text is not descriptive at all in most cases. This is because it is text coming from social networks and may contain opinions, criticism, sarcasm, humor, tags, emojis, etc. In any case, we can use the common space offered by CLIP to determine the similarity with each of the tags. In addition, we can set a confidence threshold, as we did with the images, which if it is high enough, will ensure that in the vast majority of cases it will help determine the classification of the image. To do this, using CLIP we must encode the embedding vectors of the text associated with each image and with the labels, calculating their cosine similarity. In this way, it will allow us to compare the similarity of the two texts and establish whether or not to use that value in the classification of the image.

Once this value has been obtained, we must establish a way to decide firstly, what probability threshold would be necessary to consider the text part, and secondly, how this probability vector will influence the classification. To do that, we will perform several experiments to assess whether it is possible to improve the classification. 

\begin{figure}[htp]
    \centering
    \includegraphics[width=\textwidth]{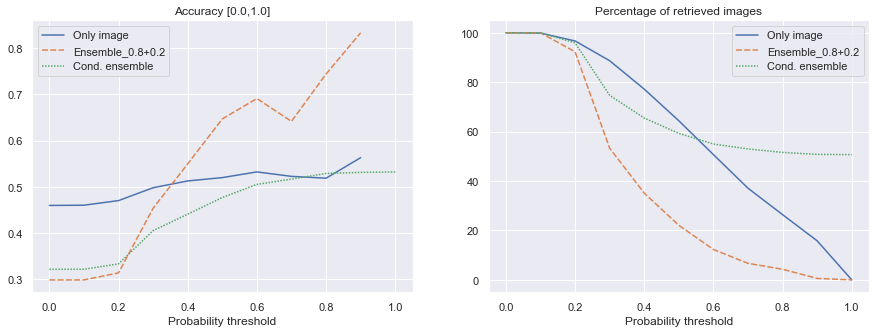}
    \caption{Comparing ensembles accuracies and number of images retrieved in the classification using text, depending on the probability threshold chosen.}
    \label{fig:plot_mix}
\end{figure}

First, we establish a basic ensemble, where we assign a weight of 0.8 to the image part and a weight of 0.2 to the textual part. In the results shown in Table \ref{fig:plot_mix} we can see that we obtain higher accuracy, starting from the threshold of 0.4, compared to that obtained with the image-only classification. On the other hand, as we increase the value of the probability threshold, we obtain a drastic decrease in the number of images recovered with this ensemble.

An example of this could be our second testing ensemble, where we set a condition of probability $< 0.6$ on only one image to complement the information obtained from the text. Nevertheless, this conditional ensemble did not improve the original ensemble.

\section{Evaluation}
\label{evaluation}

To verify that CLIP's classification process on the InstaNYC100M dataset is correct, we need to have a subset of the dataset labeled. To perform this task in a more or less efficient way, without having to resort to manual labeling of a large number of images, we have used a strategy based on a random sample of 1000 images (100 per class) classified by CLIP, as explained below.

For this validation experiment, we are choosing the 10 most frequent classes in the classification made by CLIP without setting any probability threshold, i.e., we will take the class with the highest probability, after passing through a softmax layer.  
After that, we obtain a random sample of 100 images from each class, which we will manually label as success or failure in the CLIP classification. Results are shown in Table \ref{table5}, where we can find the 10 most frequent classes in the classification of the entire dataset, together with the number of images obtained in the classification and, in the last column, the successful classification result measured as a percentage of accuracy concerning the images sectioned in the 1000 random sample. 
\begin{table}[]
\centering
\caption{Top-10 frequency classes in entire dataset and accuracy of classification in random sampling.}
\begin{tabular}{@{}lll@{}}
\toprule
Class               & Frequency in 100k      & Acc. in 1k sample \\
\midrule
skyscraper             & 7410              & 96\%                                                            \\
clothing store         & 4681              & 43\%                                                            \\
art gallery            & 4329              & 53\%                                                            \\
beauty salon           & 3879              & 24\%                                                            \\
hospital               & 3806              & 0\%                                                             \\
shower                 & 2729              & 6\%                                                             \\
crosswalk              & 2599              & 63\%                                                            \\
gift shop              & 2155              & 45\%                                                            \\
bakery shop            & 2003              & 36\%                                                            \\
bridge                 & 1763              & 94\%                                                            \\
\midrule
Average                & 35354            & 46\%       \\                                                    
\bottomrule                  
\end{tabular}
\label{table5}
\end{table}

The first conclusion we can derive from this data is that the classification accuracy is highly dependent on the label. We can observe how, for example, with the labels ``hospital'' or ``shower'' it fails to classify most of the time, while with ``skyscraper'' or ``bridge'' the accuracy is very high. An interesting fact to take into account is the average of the best probabilities of classification successes in the 1000 random sampling evaluated, 0.6579, versus the average of failures, 0.5676.  We can see that it is slightly higher and gives us an indicative range of what would be the ideal threshold [0.56-0.66].

\begin{table}[]
\centering
\caption{Successful and unsuccessful images in the classification, hit and error rates and comparative ratio depending on the probability threshold}
\begin{tabular}{@{}lllllll@{}}
\toprule
Prob thr. &\begin{tabular}[c]{@{}l@{}}Classified  \\ Images \end{tabular}  & Hits & Hit Rate & Errors & Error Rate    & \begin{tabular}[c]{@{}l@{}}Ratio \\ hit/err rates\end{tabular} \\
\midrule
0.0       & 1000   & 460      & 1          & 540      & 1          & 1                                                             \\
0.1       & 999    & 460      & 1          & 539      & 0,9981 & 1,0018                                                    \\
0.2       & 967    & 455      & 0,9891 & 512      & 0,9481 & 1,0432                                                    \\
0.3       & 887    & 442      & 0,9608 & 445      & 0,8240 & 1,1659                                                    \\
0.4       & 772    & 396      & 0,8608 & 376      & 0,6962  & 1,2363                                                    \\
0.5       & 644    & 335      & 0,7282 & 309      & 0,5722 & 1,2726                                                     \\
0.6       & 507    & 270      & 0,5869 & 237      & 0,4388 & 1,3373                                                    \\
0.7       & 371    & 194      & 0,4217 & 177      & 0,3277 & 1,2866                                                    \\
0.8       & 264    & 137      & 0,2978 & 127      & 0,2351 & 1,2663                                                    \\
0.9       & 158    & 89       & 0,1934 & 69       & 0,1277 & 1,5141                                                    \\
1.0       & 0      & 0        & 0          & 0        & 0          &                                                              \\
\bottomrule
\label{table6}
\end{tabular}
\end{table}

Table \ref{table6} shows the number of successful and unsuccessful classification images, depending on the probability threshold. In addition, we also show the accuracy in both, the hits and misses, and a comparative ratio, whose maximum value is 1.3373, which is obtained with a probability threshold of 0.6, which indicates that this value could be the optimal threshold.

\section{Conclusions and future work}
\label{future}

In this paper, we take advantage of the learning transference capability offered by CLIP to attempt to classify images in a social network. Our first proposal was based only on the vision part, to later exploit the common encoding space provided by CLIP between the language and the image by including the text associated with each picture.

We carried out a validation based on sampling techniques, obtaining metrics to evaluate the effectiveness of our work. In our experiments, we were able to obtain good results on both image classification and denoising. Although the accuracy results do not seem very high, we have to take into account that the experiments have been performed with a very noisy dataset, that the classes were very numerous, and a large set of images do not match any of these classes. Moreover, all our work has been done in zero-shot settings. By adding the texts, we obtained slight improvements in accuracy in exchange for higher discrimination of images. This can be interesting in some areas, where the quality of the sample is more important than its size.

We are still at an early stage of experimentation. Some of the promising directions are the combination of the learning transfer capability of pre-trained systems with lots of data, like CLIP, with the re-training with few-shot learning of the target dataset. In our case, trying to achieve a low-cost union between CLIP, Places dataset, and InstaNY100K (text and image) to optimize the classification is a promising research line that can be approached by several non-exclusive ways such as re-training CLIP with Places, modifying its architecture, or re-training it with a labeled subset of InstaNY100K incorporating the textual part.

Another future work is the exploitation of the associated text, not only to assist in classification, but also to obtain relevant information associated with the images, such as sentiment analysis.

\section*{Acknowledgement}
This work was funded by the University of Alicante  UAPOSTCOVID19-10  grant for ``Collecting and publishing open data for the revival of the tourism sector post-COVID-19´´ project.  We would like to thank Nvidia for their generous hardware donations that made these experiments possible.

%
% ---- Bibliography ----
%
\bibliographystyle{plain} % We choose the "plain" reference style
\bibliography{soco_2021} % Entries are in the "refs.bib"

\end{document}